\documentclass[10pt,twocolumn,letterpaper]{article}

\usepackage{cvpr}              

\usepackage{graphicx}
\usepackage{amsmath}
\usepackage{amssymb}
\usepackage{booktabs}
\usepackage{multirow}
\usepackage{xcolor}
\usepackage{color}
\usepackage[accsupp]{axessibility}  

\DeclareMathOperator{\argmax}{argmax}

\usepackage[pagebackref,breaklinks,colorlinks]{hyperref}

\usepackage[capitalize]{cleveref}
\crefname{section}{Sec.}{Secs.}
\Crefname{section}{Section}{Sections}
\Crefname{table}{Table}{Tables}
\crefname{table}{Tab.}{Tabs.}

\def\cvprPaperID{7087} 
\def\confName{CVPR}
\def\confYear{2023}

\begin{document}

\title{VisFusion: Visibility-aware Online 3D Scene Reconstruction from Videos}

\author{Huiyu Gao, Wei Mao, Miaomiao Liu\\
Australian National University\\
{\tt\small \{huiyu.gao, wei.mao, miaomiao.liu\}@anu.edu.au}}
\maketitle

\begin{abstract}
    \vspace{-0.1cm}
We propose VisFusion,~a visibility-aware online 3D scene reconstruction approach from posed monocular videos. In particular, we aim to reconstruct the scene from volumetric features.~Unlike previous reconstruction methods which aggregate features for each voxel from input views without considering its visibility, we aim to improve the feature fusion by explicitly inferring its visibility from a similarity matrix, computed from its projected features in each image pair.~Following previous works, our model is a coarse-to-fine pipeline including a volume sparsification process. Different from their works which sparsify voxels globally with a fixed occupancy threshold, we perform the sparsification on a local feature volume along each visual ray to preserve at least one voxel per ray for more fine details. The sparse local volume is then fused with a global one for online reconstruction.~We further propose to predict TSDF in a coarse-to-fine manner by learning its residuals across scales leading to better TSDF predictions.~Experimental results on benchmarks show that our method can achieve superior performance with more scene details. Code is available at: \url{https://github.com/huiyu-gao/VisFusion}
\vspace{-0.1cm}
\end{abstract}

\section{Introduction}
\label{sec:intro}

3D scene reconstruction from RGB videos is a critical task in 3D computer vision, which finds its broad applications in augmented reality (AR), robot navigation and human-robot interaction. These applications require accurate, complete and real-time 3D reconstruction of scenes. While state-of-the-art SLAM systems~\cite{campos2021orb, von2018direct} can track the camera motion accurately by leveraging both visual and inertial measurements in an unknown environment, the reconstructed map from a SLAM system only contains sparse point clouds such that dense reconstruction from monocular videos remains as a challenging problem.

Many previous methods~\cite{murez2020atlas,bleyer2011patchmatch} assume the observation of the whole video sequence for the reconstruction, which is not practical for online applications like VR games. In this paper, we follow~\cite{sun2021neuralrecon} to propose an online 3D reconstruction method. Given input images, most earlier 3D reconstruction methods~\cite{schops20153d, yang2017real} adopt a two-stage pipeline, which first estimates the depth map for each keyframe based on multi-view stereo (MVS) algorithms~\cite{wang2018mvdepthnet, liu2019neural, hou2019multi, valentin2018depth} and then fuses the estimated depth maps into a Truncated Signed Distance Function (TSDF) volume~\cite{newcombe2011kinectfusion}. The Marching Cubes algorithm~\cite{lorensen1987marching} is then used to extract the 3D mesh. However, those two-stage pipelines struggle to produce globally coherent reconstruction since each depth map is estimated separately~\cite{sun2021neuralrecon}, especially for low texture regions like walls whose depth values are extremely hard to estimate with only several local views. To address this, more recent works~\cite{sun2021neuralrecon, bozic2021transformerfusion, xie2022planarrecon} propose to fuse image features into a global 3D volume and directly regress TSDF~\cite{sun2021neuralrecon, xie2022planarrecon} or occupancy~\cite{bozic2021transformerfusion} given the feature volume. Such strategy allows for an end-to-end global surface reconstruction. 

The problem of occlusion naturally arises for global feature fusion. Previous methods~\cite{sun2021neuralrecon,bozic2021transformerfusion} either completely ignore it by simply averaging the multi-view features~\cite{sun2021neuralrecon} for each voxel or implicitly model the visibility via the attention mechanism~\cite{bozic2021transformerfusion}. However, without explicit supervision, such attention cannot guarantee to encode the correct visibility. In this paper, we thus propose to explicitly predict the visibility weights of all views for each voxel with ground truth supervision. In addition, voxels will be considered~\emph{visible} in at least one view in~\cite{bozic2021transformerfusion} due to the normalization of the attention mechanism, while in our method, empty voxels and fully occluded voxels are~\emph{invisible} in any view to avoid introducing noises. Specifically, given a fragment of a video sequence observing the same 3D region, we first project each 3D voxel onto different view images to obtain 2D features. We then compute the pair-wise similarities of these features. Since features of the same occupied voxel are often similar across views, such similarity map naturally encodes the information of whether a 3D voxel is visible at a particular camera view or not  (see~\cref{fig:similarity}). We thus use this similarity map to predict visibility weights.

For volumetric-based methods, it is common practice to adopt a coarse-to-fine pipeline~\cite{murez2020atlas,sun2021neuralrecon,bozic2021transformerfusion,stier2021vortx}. One of its key steps is voxel sparsification which eliminates empty voxels at coarse level for better performance and smaller memory consumption. To the best of our knowledge, previous methods~\cite{murez2020atlas,sun2021neuralrecon,bozic2021transformerfusion,stier2021vortx} propose to globally sparsify the volume by removing voxels whose occupancy probabilities are lower than a predefined threshold. However, such fixed threshold tends to sparsify more voxels than necessary, especially to remove voxels covering thin structures such as chair legs. At coarse level where the thin structure only occupies a small portion of the voxel, the features of such thin structure are likely ignored leading to low occupancy probability prediction and resulting in the removal of such voxel. However, such voxel should rank highly, based on the occupancy probability, among voxels along the visual ray defined by the pixel observing this thin structure. Inspired by this, we introduce a novel ray-based sparsification process. In particular, for any image, we first cast a ray from every pixel to get the voxels this ray passes. For each ray, we then keep voxels with top occupancy scores to next level. Unlike previous works~\cite{murez2020atlas,sun2021neuralrecon,bozic2021transformerfusion,stier2021vortx} that sparsify the global volume, our ray-based sparsification is performed on local 3D volume. Our ray-based sparsifying strategy allows us to retain more surface voxels to the next level leading to a more complete reconstruction. 

Furthermore, previous coarse-to-fine methods~\cite{murez2020atlas,sun2021neuralrecon,bozic2021transformerfusion,stier2021vortx} directly regress the TSDF at each level discarding the relationships between the TSDF predicted at coarse and that at fine level. In our method, at each fine level, we aim to predict a residual between the TSDF volume upsampled from the coarser level and that of the fine level, which is shown to be more accurate in TSDF estimation.

In summary, our contributions are 
(i) a visibility-aware feature fusion module which explicitly predicts visibility weights used for feature aggregation for voxels; 
(ii) a ray-based voxel sparsifying algorithm which leads to the reconstruction of more scene structure details.
(iii) an easier way of TSDF regression by learning the residual to the upsampled coarse TSDF volume for improved TSDF estimation. Our model outperforms the existing online feature fusion based methods. 

\section{Related Work}
\label{sec:related_work}

\begin{figure*}
  \centering
  \includegraphics[width=0.96\linewidth]{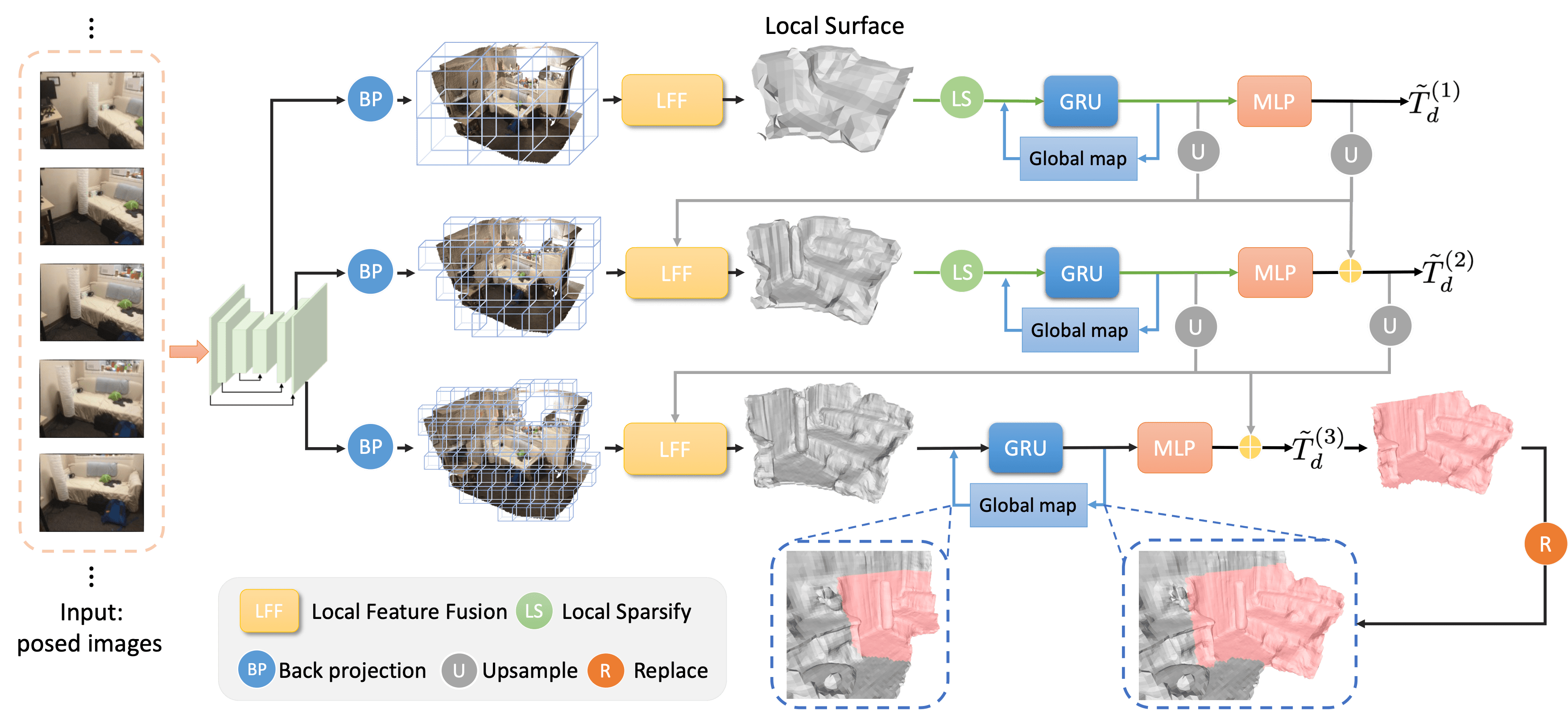}
  \vspace{-0.2cm}
  \caption{\textbf{Overview of our coarse-to-fine pipeline.} Given a fragment of a video, we first construct 3D feature volumes of different resolutions. The feature of each voxel is obtained by projecting it back to every camera view. The features from different camera views are then fused via the predicted visibility (local feature fusion). We then extract the local occupancy and TSDF from the fused feature followed by a ray-based sparsification process to remove empty voxels. The sparse local feature volume is finally fused to global via GRU and used to produce the final TSDF. The global feature and final TSDF of coarse level are further upsampled and fed to the next level for refinement.}
  \label{fig:overview}
  \vspace{-0.04cm}
\end{figure*}

\noindent\textbf{Multi-view depth estimation.} 
COLMAP~\cite{schonberger2016pixelwise} is one of the most popular methods for Multi-View Stereo which jointly estimate the depth and normal with photometric and geometric consistency. Although it is robust and accurate, COLMAP has difficulty in densely reconstructing areas without distinctive features. Deep learning-based methods~\cite{wang2018mvdepthnet, hou2019multi, im2019dpsnet} try to relax the dependency of the photo-consistency assumption with data-driven priors. Although some recent studies\cite{yao2018mvsnet, im2019dpsnet, duzceker2021deepvideomvs, hou2019multi, liu2019neural} show that constructing 3D cost volumes followed by 3D CNNs for the cost volume processing could achieve promising results, they have large gpu memory cost and struggle to achieve real-time inference speed. Recently, SimpleRecon~\cite{sayed2022simplerecon} utilizes image priors and integrates the geometric metadata into the cost volume without using 3D convolutions. The scene geometry is then obtained by a depth fusion approach, such as volumetric TSDF fusion~\cite{curless1996volumetric}, which fuses the estimated depth maps into geometric 3D representation. While the incremental volumetric TSDF fusion method proposed by KinectFusion~\cite{newcombe2011kinectfusion} is widely applied due to its simplicity and effectiveness, such two-stage approach suffers from computational redundancy and causes artifacts in the 3D reconstruction as the depth maps are estimated separately.

\noindent\textbf{End-to-end 3D scene reconstruction.}
To achieve coherent reconstruction, several volumetric approaches are proposed to directly infer the 3D geometry from volumetric features in an end-to-end manner. SurfaceNet~\cite{ji2017surfacenet} is the first volumetric method to predict volumetric surface occupancy from two input images with a 3D convolutional network. Atlas~\cite{murez2020atlas} extends this work to multi-view setup by averaging the features back projected from all input images in the sequence and directly regressing the global TSDF volume of the scene. VolumeFusion~\cite{choe2021volumefusion} and 3DVNet~\cite{rich20213dvnet} further improve Atlas by using the estimated depth maps of input views as intermediate representations whose training is supervised by ground truth depth supervision.~VoRTX~\cite{stier2021vortx} adopts the transformer architecture~\cite{vaswani2017attention} for multi-view feature fusion. While improving the reconstruction quality, these methods~\cite{murez2020atlas, choe2021volumefusion, rich20213dvnet, stier2021vortx} can only work in an offline fashion, requiring entire image sequences as input, thus cannot achieve online reconstruction.

For online end-to-end 3D reconstruction methods, NeuralRecon~\cite{sun2021neuralrecon} proposes the first real-time framework for dense reconstruction from posed monocular videos. It converts Atlas~\cite{murez2020atlas} to an incremental reconstruction system by averaging features only within local video fragments and fusing across fragments using a 3D convolutional variant of Gated Recurrent Unit (GRU)~\cite{chung2014empirical} module. However, simply fusing image features by averaging in local fragments does not effectively model the visibility and solve occlusion issues. In addition, although NeuralRecon employs a coarse-to-fine approach to reduce memory consumption, it uses a global threshold to sparsify voxels between layers, resulting in losing fine structures. Similar to~\cite{stier2021vortx}, TransformerFusion~\cite{bozic2021transformerfusion} also leverages the transformer architecture~\cite{vaswani2017attention} to fuse features in the global voxel space. It performs in an online manner by saving features from multiple views and dropping saved features with the lowest attention weight. However, this design makes TransformerFusion suffer from high computing costs and storage consumption. Recently, PlanarRecon~\cite{xie2022planarrecon} proposes an incremental framework for 3D plane detection and fusion, which mainly focuses on planar region reconstruction. In contrast, we aim for the reconstruction of the scene structure in general.

\noindent\textbf{Residual learning and refinement.}
In MVS task, many depth estimation works~\cite{chen2019point, gu2020cascade, yang2020cost} predict a residual to an initial depth prediction and refine it gradually across multi-scales. Inspired by such strategy, we propose a cascaded TSDF learning framework that gradually learns the TSDF residual to the one learned from a coarser level and reconstructed from previous fragments. By doing so, our model is able to achieve better reconstruction performance.
\section{Method}
\label{sec:method}

\begin{figure*}
  \centering
  \includegraphics[width=0.96\linewidth]{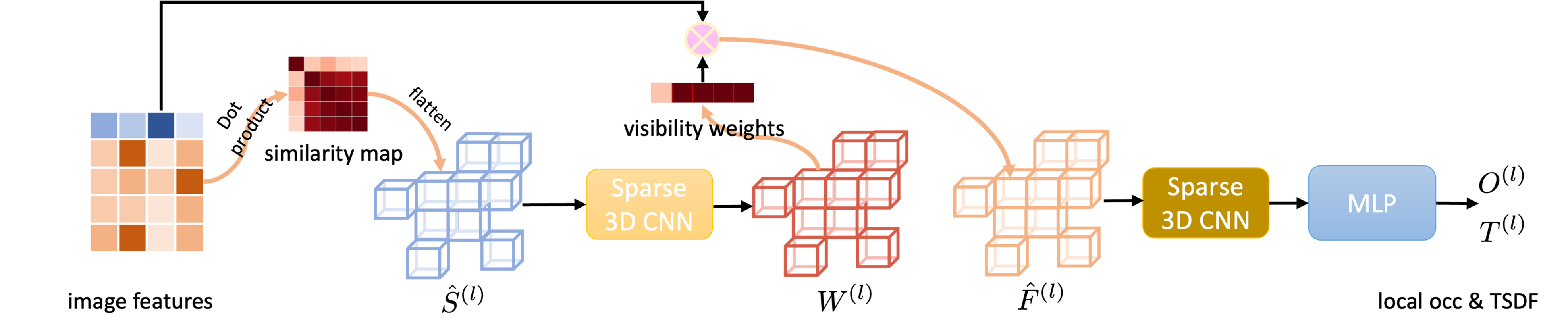}
  \vspace{-0.4cm}
  \caption{\textbf{Local feature fusion.} For each voxel, given the $N$ features extracted from different camera views, we first compute their pairwise cosine similarities. Such similarity map is then flattened as a vector. The 3D volume of such vectors is processed by a 3D CNN to produce visibility weights $\mathbf{W}$, which is used to aggregate the back projected features. With another 3D CNN followed by one MLP layer, we obtain the local occupancy and TSDF for our ray-based sparsification. }
  \vspace{-0.1cm}
  \label{fig:local_fusion}
\end{figure*}
Let us now introduce our approach to online 3D scene reconstruction from posed monocular videos.  Given a sequence of monocular images $\{\mathbf{I}_i\}$ of a scene, where $\mathbf{I}_i \in \mathbb{R}^{H \times W \times 3}$ represents the $i^{\text{th}}$ RGB image in the video sequence, with their corresponding camera intrinsics, rotation  matrices, and translation vectors $\{\mathbf{K}_i\in\mathbb{R}^{3\times 3}, \mathbf{R}_i\in\mathbb{R}^{3\times 3}, \mathbf{t}_i\in\mathbb{R}^{3}\}$ estimated by a SLAM system, our method incrementally reconstructs the dense 3D geometry of the scene.~To achieve online reconstruction, following~\cite{sun2021neuralrecon, hou2019multi}, we sequentially select suitable keyframes. A new incoming frame is selected as a keyframe if the camera motion is greater than a predefined threshold~\cite{sun2021neuralrecon}. We then split the keyframe stream into $M$ non-overlapping fragments $\mathcal{F}_t = \{\mathbf{I}_{i} \}_{i=N(t-1) + 1}^{N t}, t = \{1, 2, \dots, M\}$, each of which consists of $N$ consecutive keyframes. The view-frustum of each keyframe image is computed with a fixed max depth range $d_{\text{max}}$. Same as~\cite{sun2021neuralrecon}, we only reconstruct the cubic-shaped region that encloses the view-frustums of all images within this fragment. Such region is called fragment bounding volume (FBV)~\cite{sun2021neuralrecon}.

An overview of our framework is shown in \cref{fig:overview}. Our method predicts the TSDF representation of the scene, in a coarse-to-fine manner. It consists of three modules, namely, the local feature fusion module, the local sparsification module, and the global feature fusion module. Below, we introduce these modules in detail.

\subsection{Local Feature Fusion}\label{sec:localfusion}
Here we discuss our proposed local feature fusion module at level $l$ of our coarse-to-fine framework. Given $N$ images in current fragment $\mathcal{F}_t$, our local feature fusion module aims to aggregate image features according to the visibility of voxels in the corresponding FBV. In particular, the local FBV is represented as a 3D voxel grid of $D^{(l)}$ voxels $\mathbf{V}^{(l)}\in \mathbb{R}^{D^{(l)}\times 3}$. Note that, the FBVs at different levels have different resolutions thus different numbers of voxels~\footnote{At the first (coarsest) level, the local volume will be a full grid while for the rest, the volumes will be a small subset of the full grid upsampled from the sparsified ones of the previous level.}. We follow~\cite{sun2021neuralrecon} to use a variant of MnasNet~\cite{tan2019mnasnet} to extract 2D features from input images. Each volume is then projected to different camera views to obtain an initial local feature volume $\mathbf{F}^{(l)}\in\mathbb{R}^{D^{(l)}\times N\times C^{(l)}}$ as,
\begin{equation}
    \mathbf{F}^{(l)}_{dn} = \mathbf{H}_{n}^{(l)}(\Pi(\mathbf{K}^{(l)}_{n},\mathbf{R}_{n}\mathbf{V}_{d}^{(l)}+\mathbf{t}_{n}))\;,
\end{equation}
where $\mathbf{F}^{(l)}_{dn}\in\mathbb{R}^{C^{(l)}}$ is the feature vector of voxel $\mathbf{V}_{d}^{(l)}$ obtained from $n^{\text{th}}$ image feature map $\mathbf{H}_{n}^{(l)}\in \mathbb{R}^{H^{(l)}\times W^{(l)} \times C^{(l)}}$. With a little abuse of notation, we also use $\mathbf{H}_{n}^{(l)}(\cdot)$ to represent 2D feature interpolation. $\Pi(\cdot,\cdot)$ is the camera perspective projection. Note that, we use different resolutions of 2D image features for different levels and the camera intrinsics are also changed accordingly. For each voxel, given the features from $N$ different views $\{\mathbf{F}^{(l)}_{dn}\}_{n=1}^N$, our goal now is to fuse these local features into one. Unlike previous scene reconstruction methods~\cite{murez2020atlas, sun2021neuralrecon} that fuse those features by simply averaging them, we propose to directly predict the visibility weights for feature fusion.

\begin{figure}[!t]
  \centering
  \begin{subfigure}{0.32\linewidth}
    \includegraphics[width=1\linewidth]{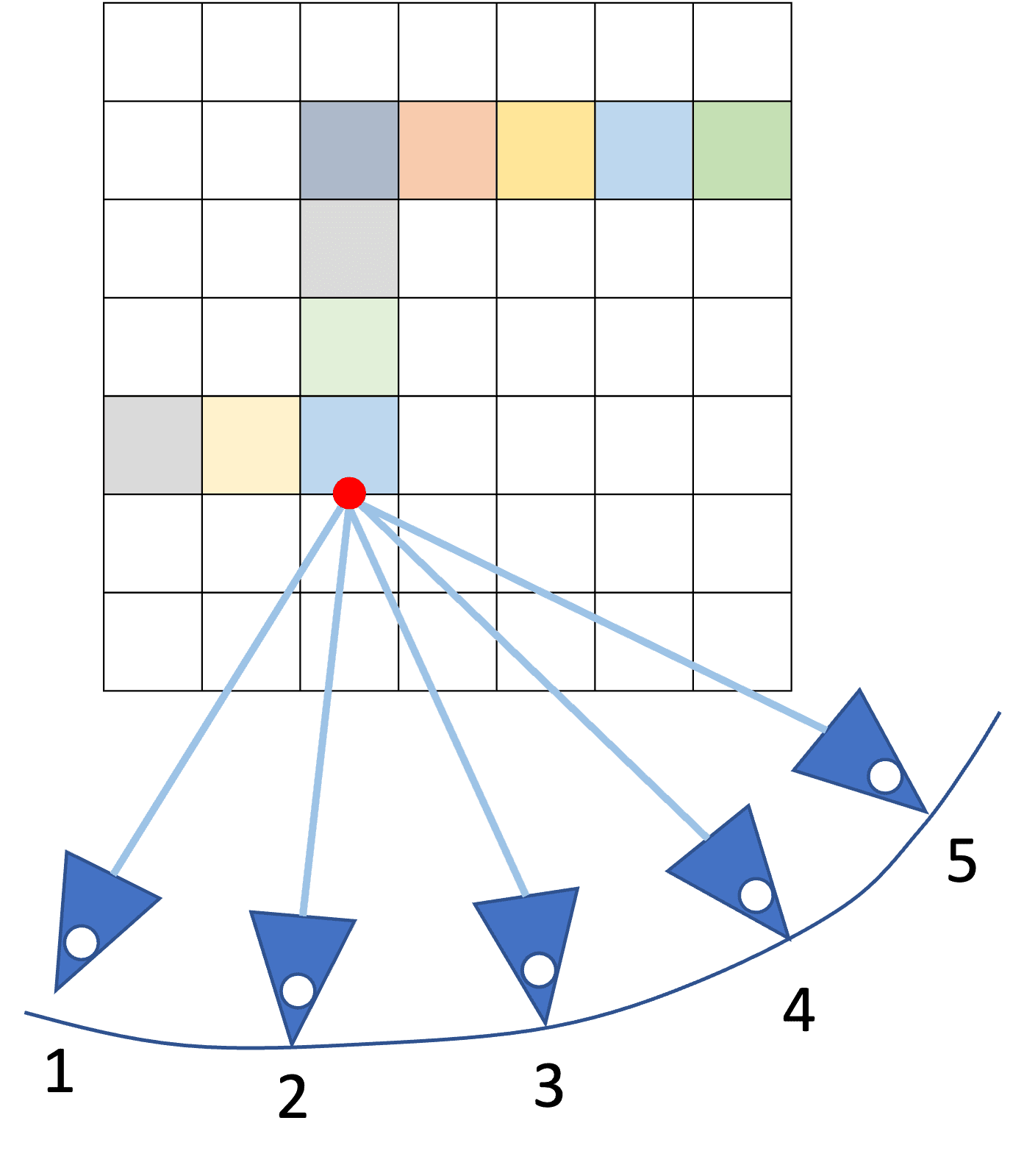}
    \caption{}
    \label{fig:similarity-a}
  \end{subfigure}
  \hfill
  \begin{subfigure}{0.32\linewidth}
    \includegraphics[width=1\linewidth]{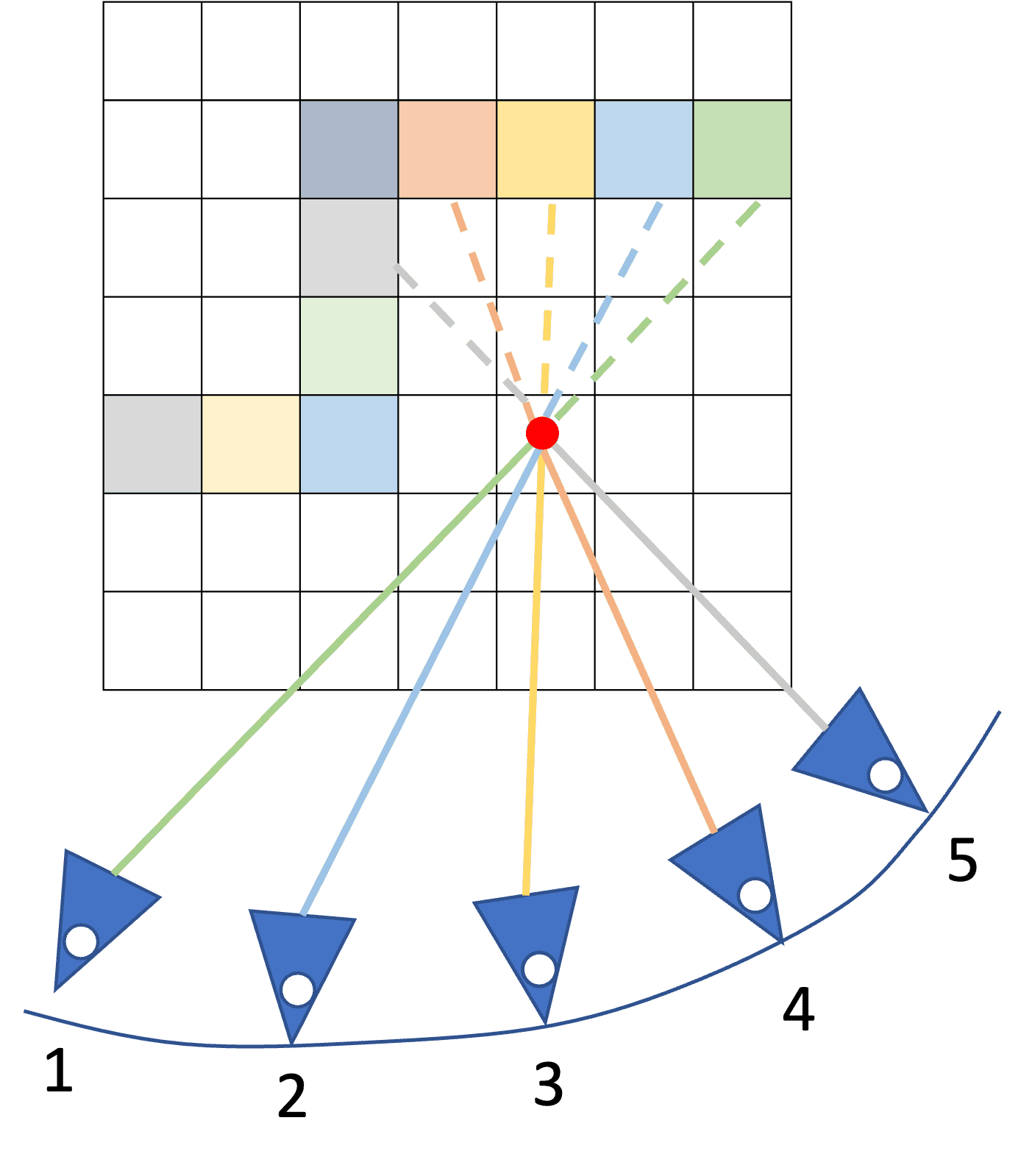}
    \caption{}
    \label{fig:similarity-b}
  \end{subfigure}
    \begin{subfigure}{0.32\linewidth}
    \includegraphics[width=1\linewidth]{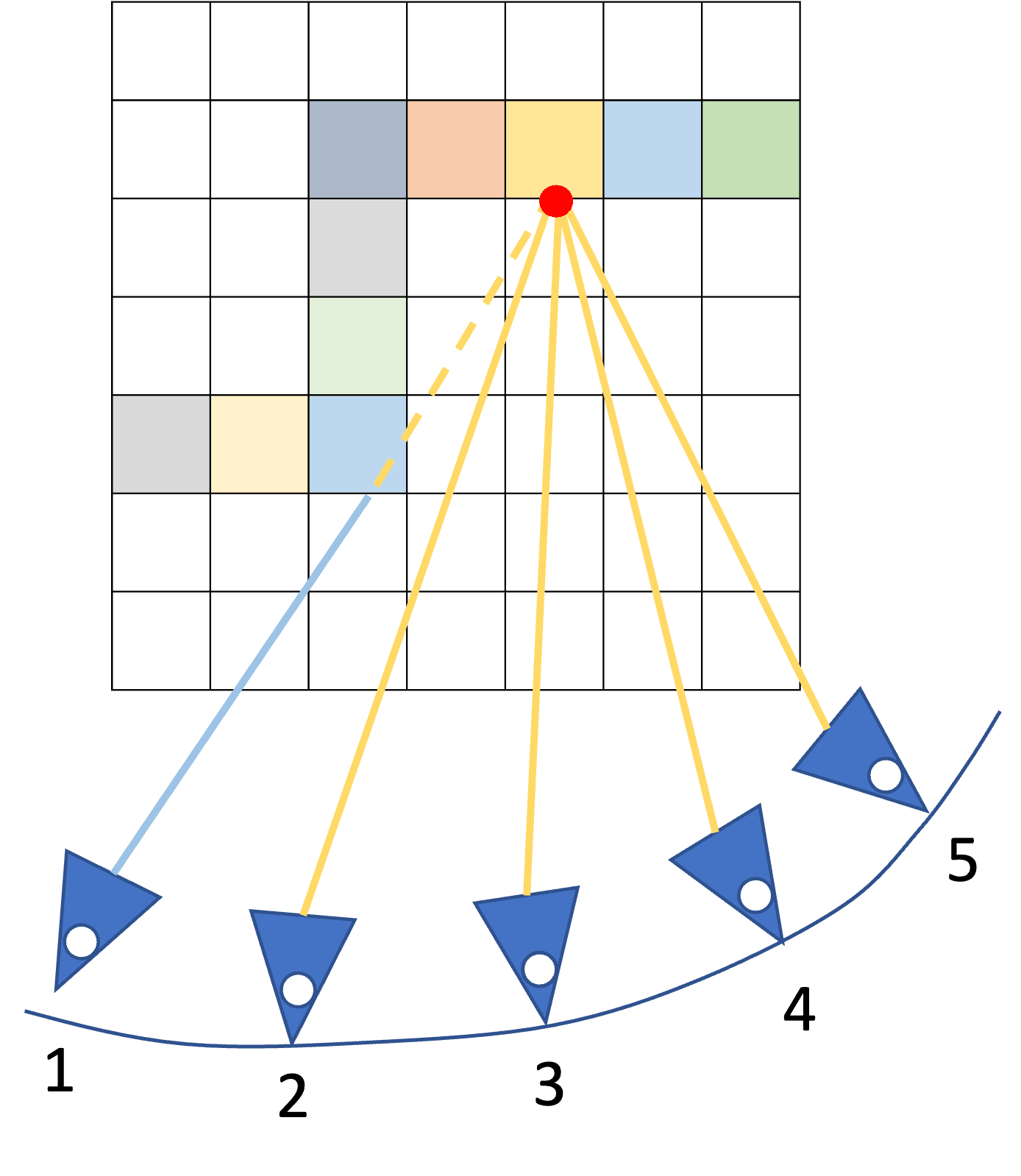}
    \caption{}
    \label{fig:similarity-c}
  \end{subfigure}
    \vspace{-0.1cm}
    \begin{subfigure}{0.32\linewidth}
    \includegraphics[width=1\linewidth]{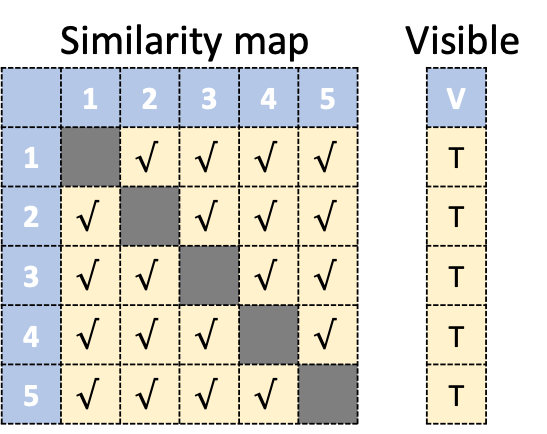}
    \caption{}
    \label{fig:similarity-d}
  \end{subfigure}
  \hfill
  \begin{subfigure}{0.32\linewidth}
    \includegraphics[width=1\linewidth]{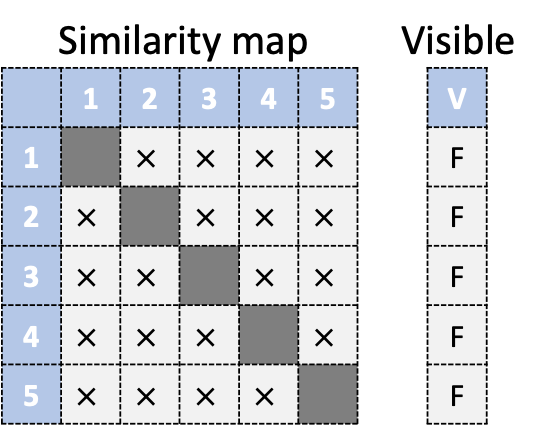}
    \caption{}
    \label{fig:similarity-e}
  \end{subfigure}
    \begin{subfigure}{0.32\linewidth}
    \includegraphics[width=1\linewidth]{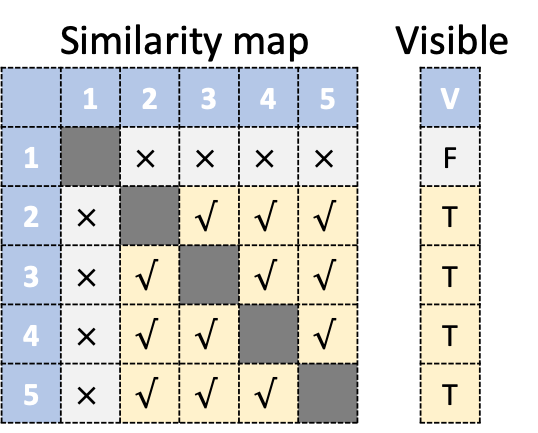}
    \caption{}
    \label{fig:similarity-f}
  \end{subfigure}
  \vspace{-0.1cm}
  \caption{\textbf{Different kinds of voxels and their similarity maps.} $\surd$ denotes relatively similar and $\times$ denotes relatively dissimilar. T represents True and F represents False. (a) Surface voxel (b) Empty voxel (c) Surface voxel with occlusion (d), (e), (f) Similarity map and visibility definition corresponding to (a), (b), (c).}
  \label{fig:similarity}
\end{figure}

Let us first define the visibility of a voxel. A voxel is considered to be occupied when its distance to the nearest surface is less than the truncation distance $\lambda$ of the TSDF. A voxel is visible to a view if it is occupied and not occluded by other occupied voxels. This differs from the visibility defined in~\cite{stier2021vortx, liu2022neural} where empty voxels are also treated as visible as long as they are not occluded~\footnote{Based on the definition of visibility in VoRTX~\cite{stier2021vortx} and NeuRay~\cite{liu2022neural}, besides surface voxels (points), all empty voxels (points) along the ray before it hits the surface are also defined as~\emph{visible}. By contrast, we define those empty and occluded voxels as~\emph{invisible} to avoid introducing noises to our feature volume.}. As illustrated in~\cref{fig:local_fusion}, we proposed to predict the visibility from the pairwise feature similarities. In particular, given the features of a voxel $\mathbf{F}^{(l)}_{dn}$, we compute the cosine similarity between features from every two views as, 
\begin{equation}
\mathbf{S}^{(l)}_{dmn}=\frac{\mathbf{F}^{(l)^T}_{dm}\mathbf{F}^{(l)}_{dn}}{\|\mathbf{F}^{(l)}_{dm}\|_2\|\mathbf{F}^{(l)}_{dn}\|_2}\;,
\end{equation}
where $\mathbf{S}^{(l)}_{d}\in\mathbb{R}^{N\times N}$ is the similarity map of this voxel and $m,n\in\{1,2,\cdots,N\}$. As illustrated in \cref{fig:similarity}, such similarity maps provide important geometry heuristics for predicting the visibility of each voxel in the local fragment. We then eliminate the diagonal entries which are always ones and flatten the map to obtain a long similarity feature denoted as $\hat{\mathbf{S}}^{(l)}_{d}\in\mathbb{R}^{N(N-1)}$. For the whole FBV, we then have a similarity volume $\hat{\mathbf{S}}^{(l)}\in\mathbb{R}^{D^{(l)}\times N(N-1)}$. It is fed into a 3D CNN to obtain visibility weights $\hat{\mathbf{W}}^{(l)}\in\mathbb{R}^{D^{(l)}\times N}$. For each voxel, there are $N$ weights that will be used to fuse the local feature volume $\mathbf{F}^{(l)}$ as,
\begin{equation}
\hat{\mathbf{F}}^{(l)}_{d}=\sum_{n=1}^{N}\hat{\mathbf{W}}^{(l)}_{dn}\mathbf{F}^{(l)}_{dn}\;,
\end{equation}
where $\hat{\mathbf{F}}^{(l)}\in\mathbb{R}^{D^{(l)}\times C^{(l)}}$ is the fused local feature volume. Another 3D CNN followed by a linear layer is then used to generate local occupancy $\hat{\mathbf{O}}^{(l)}\in\mathbb{R}^{D^{(l)}}$ and local TSDF $\hat{\mathbf{T}}^{(l)}\in\mathbb{R}^{D^{(l)}}$. 

\noindent\textbf{Training loss.} The training losses for the visibility, local occupancy and local TSDF are defined as follows.
\begin{equation}
    \mathcal{L}_{w}^{(l)}= \frac{1}{ND^{(l)}}\sum_{d=1}^{D^{(l)}}\sum_{n=1}^{N}(\hat{\mathbf{W}}^{(l)}_{dn}-\frac{\mathbf{W}^{(l)}_{dn}}{\sum_{m=1}^{N}\mathbf{W}^{(l)}_{dm}})^2\;,
\end{equation}
where $\mathbf{W}^{(l)}_{dn}\in\{0,1\}$ is the ground truth visibility which indicates whether the $d^{\text{th}}$ voxel is visible in $n^{\text{th}}$ view and $\hat{\mathbf{W}}^{(l)}_{dn}\in [0,1]$. Note that, here we directly train our model to output the normalized visibility.
\begin{equation}
    \mathcal{L}_{o}^{(l)}= \frac{1}{D^{(l)}}\sum_{d=1}^{D^{(l)}}\text{BCE}(\hat{\mathbf{O}}^{(l)}_{d},\mathbf{O}^{(l)}_{d})\;,
    \label{eq:loss_occ_local}
\end{equation}
where $\text{BCE}(\cdot,\cdot)$ is the binary cross entropy; $\mathbf{O}^{(l)}_{d}\in\{0,1\}$ is the ground truth occupancy and $\hat{\mathbf{O}}^{(l)}_{d}\in[0,1]$.
\begin{equation}
    \mathcal{L}_{t}^{(l)}=\frac{1}{D^{(l)}}\sum_{d=1}^{D^{(l)}}|\ell(\hat{\mathbf{T}}^{(l)}_{d})-\ell(\mathbf{T}^{(l)}_{d})|\;,
    \label{eq:loss_sdf_local}
\end{equation}
where $\ell(x)=\text{sgn}(x)\log(|x|+1)$ is the log scale function as used in~\cite{sun2021neuralrecon} and $\text{sgn}(\cdot)$ is the sign function; $\mathbf{T}^{(l)}_{d}\in[-1,1]$ is the ground truth TSDF and $\hat{\mathbf{T}}^{(l)}_{d}\in[-1,1]$.

With the local occupancy at level $l$, we would like to discard the empty voxels and only upsample the remaining ones to the next level $l+1$. Existing methods~\cite{murez2020atlas,sun2021neuralrecon} that rely on a fixed threshold tend to sparsify more voxels than necessary leading to incomplete reconstruction especially for thin structures. In the next section, we introduce our ray-based local sparsification approach to address this.

\vspace{-0.2cm}

\subsection{Ray-based Local Sparsification}\label{sec:localsparsify}
\begin{figure}[!t]
  \centering
  \includegraphics[width=0.85\linewidth]{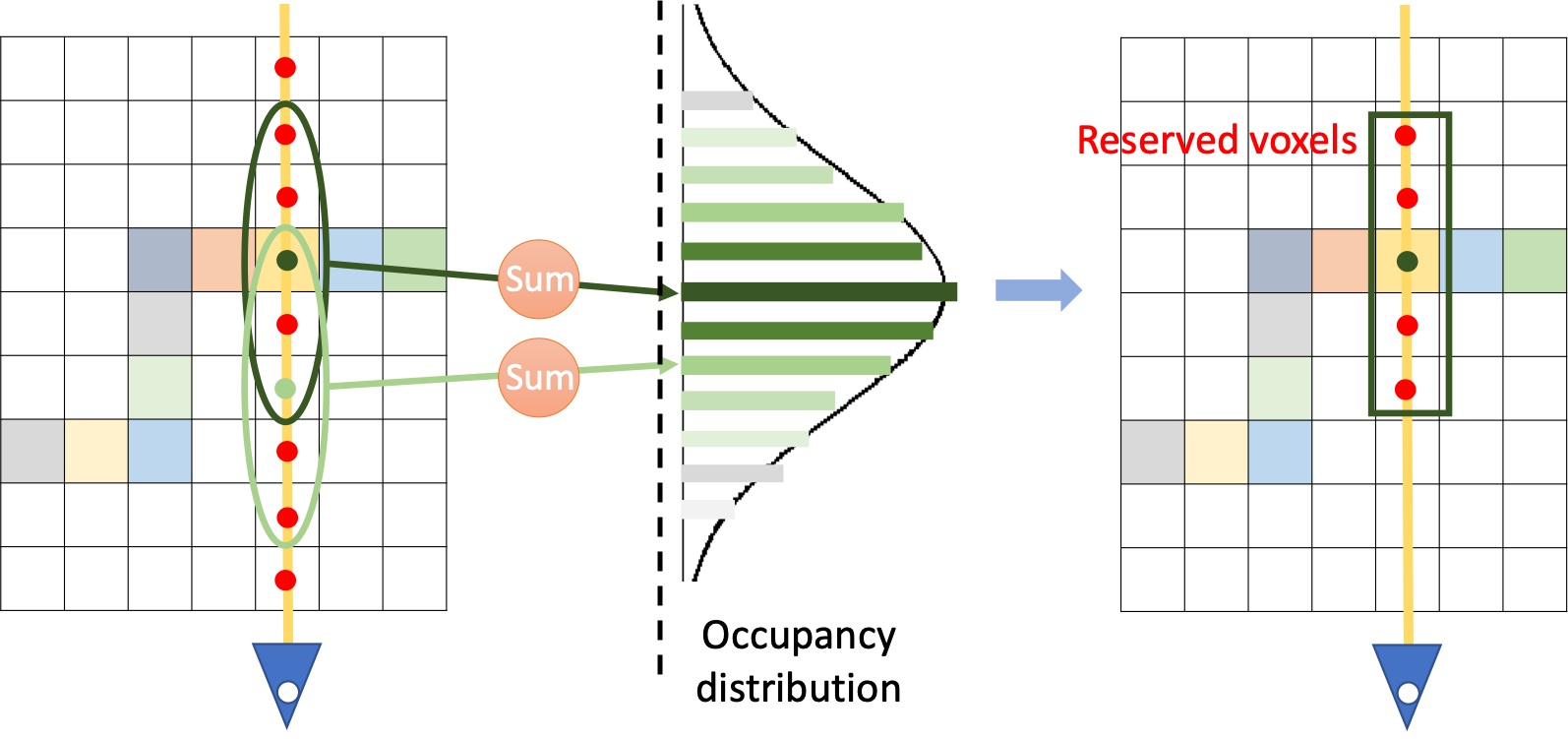}
  \vspace{-0.2cm}
  \caption{\textbf{Ray-based local sparsification.} Given the voxels passed by a visual ray and their local occupancies, we obtain a set of sliding windows, each of which covers several consecutive voxels. The sliding window with the highest sum of occupancies computed from covered voxels is reserved for the next level.}
  \vspace{-0.1cm}
  \label{fig:sparsify}
\end{figure}

Given the camera parameters $\{\mathbf{K}_i^{(l)}, \mathbf{R}_i, \mathbf{t}_i\}_{i=N(t-1)+1}^{Nt}$ of images in current fragment $\mathcal{F}_{t}$, we first cast a ray from each pixel in these images. Given a ray, let us denote the voxels that are passed by this ray as $\{\mathbf{v}_i\}_{i=1}^{R}$ where $\mathbf{v}_i\in\mathbb{R}^{3}$ is the 3D coordinates of the voxel centre. These voxels are organized in ascending order by their depth to the corresponding camera view as shown in~\cref{fig:sparsify}. Their occupancies are represented as $\{\hat{o}_i\}_{i=1}^{R}$ where $\hat{o}_i\in[0,1]$ is obtained from the predicted local occupancy volume using the voxel coordinate $\mathbf{v}_i$.

For each ray, one can then simply choose voxels with top-k occupancies. However, such strategy is not robust enough and as will be shown in our experiments, is not optimal. Alternatively, we propose to use a sliding window along the ray and rank each sliding window according to the sum of all the voxel occupancies in it. Specifically, let us assume the size of such sliding window are $K$ and $K<R$, the set of sliding windows can then be represented as $\{\mathcal{W}_i\}_{i=1}^{R-K}$ where $\mathcal{W}_i=\{\mathbf{v}_j\}_{j=i}^{i+K-1}$. The sum of occupancies for $i^{\text{th}}$ sliding window is, $\hat{u}_i=\sum_{j=i}^{i+K-1}\hat{o}_i$. The sliding window with the highest sum of occupancy scores is then selected as $i^* = \argmax_{i} \hat{u}_i$.
By doing so, our sparsification strategy adaptively selects the most likely occupied voxels for every ray which is better than only relying on a fixed threshold. We use the same criteria to select sliding windows for each ray in every input image. A voxel is sparsified only if it is not in any of these selected sliding windows. Note that, our ray-based sparsification is performed at the local volume instead of the global one. This is to avoid the situation when a ray not only passes the current surface but also existing surfaces in the global volume. In this case, there are potentially multiple sliding windows with high sum of occupancies.

\subsection{Global Feature Fusion}\label{sec:globalfusion}
Our model maintains a global feature volume which stores the information from all historically reconstructed fragments~\cite{sun2021neuralrecon}. The stored global feature volume may overlap with the FBV of the current fragment. We follow~\cite{sun2021neuralrecon} to use a 3D convolutional variant of Gated Recurrent Unit (GRU)~\cite{chung2014empirical} to fuse the local feature volume to global. Specifically, after local sparsification, we have a set of local features denoted as $\mathbf{L}^{(l)}$ and the corresponding global features as $\mathbf{G}^{(l)}$. Note that, for the overlapping local volume, its corresponding global feature is directly obtained from the stored volume. For the non-overlapping ones, the global feature will be initialized as zeros. The fused global feature is then updated as
\begin{equation}
    \Tilde{\mathbf{G}}^{(l)}=\text{GRU}(\mathbf{L}^{(l)},\mathbf{G}^{(l)})\;,
\end{equation}
where $\mathbf{G}^{(l)}$ and $\mathbf{L}^{(l)}$ are used as the hidden state and input to the GRU, respectively. The fused $\Tilde{\mathbf{G}}^{(l)}$ are then used for global TSDF prediction via a single layer MLP. 

Unlike existing coarse-to-fine works~\cite{sun2021neuralrecon,stier2021vortx} that directly regress the TSDF, we propose an easier way by taking advantage of the predicted TSDF at previous coarse level. In particular, given the global TSDF volume estimated from the above fused global feature $\Tilde{\mathbf{T}}^{(l-1)}$ at level $l-1$, we first upsample it to match the resolution of current level as $\Tilde{\mathbf{T}}^{(l-1)\uparrow}$. The output of current level is treated as a residual $\Delta\Tilde{\mathbf{T}}^{(l)}$ and the final global TSDF at level $l$ is,
\begin{equation}
    \Tilde{\mathbf{T}}^{(l)} = \Tilde{\mathbf{T}}^{(l-1)\uparrow} + \Delta\Tilde{\mathbf{T}}^{(l)}\;.
\end{equation}
As shown in our ablation study in next section, such strategy leads to better performance. Similar to~\cref{eq:loss_occ_local} and~\cref{eq:loss_sdf_local}, we can then compute the global occupancy and TSDF loss as $\mathcal{L}_{O}^{(l)}$ and $\mathcal{L}_{T}^{(l)}$.

\noindent\textbf{Overall training loss.} The overall training loss for the whole model is then,
\begin{equation}
    \mathcal{L} = \sum_{l=1}^{3} \omega^{(l)}(\mathcal{L}_{w}^{(l)} + \mathcal{L}_{o}^{(l)} + \mathcal{L}_{t}^{(l)} +\mathcal{L}_{O}^{(l)} +\mathcal{L}_{T}^{(l)})\;,
\end{equation}
where $\omega^{(l)}$ is the loss weight at level $l$.

\subsection{Implementation Details}
Our network is trained using Adam optimizer with batch size of 2 on an Nvidia RTX 3090 GPU for 50 epochs. The loss weights $\omega^{(1)}$, $\omega ^{(2)}$, $\omega^{(3)}$ are set as 1, 0.8, 0.64. We choose the checkpoint of the last training epoch for evaluation. Following~\cite{sun2021neuralrecon}, we use torchsparse \cite{tang2020searching} for 3D sparse convolution and initialize the feature extraction backbone, a variant of MnasNet~\cite{tan2019mnasnet}, with the pretrained weights from ImageNet. In each local fragment, the number of key frame images $N$ is set to 9. We use 3 coarse-to-fine layers and set the voxel size for each layer as 16 cm, 8 cm, and 4 cm respectively. The TSDF truncation distance $\lambda$ is set as three times the voxel size for each layer. The size of the sliding window $K$ used in our local sparsification module is 9 across all layers with different resolutions.
\section{Experiments}
\label{sec:experiments}

\begin{table}[!t]
  \centering
  \resizebox{\linewidth}{!}{
  \begin{tabular}{@{}ccccccccc@{}}
    \toprule
    & & Method & Acc $\downarrow$ & Comp $\downarrow$ & Chamfer $\downarrow$ & Prec $\uparrow$ & Recall $\uparrow$ & F-score $\uparrow$\\
    \midrule
    \multirow{7}{*}{\rotatebox[origin=c]{90}{Depth Fusion}}
    &&COLMAP~\cite{schonberger2016pixelwise} & 13.5 & \textbf{\textcolor{blue}{6.9}} & 10.2 & 0.505 & \textbf{\textcolor{blue}{0.634}} & 0.558\\
    &&MVDNet~\cite{wang2018mvdepthnet} & 20.5 & 8.4 & 14.5 & 0.231 & 0.473 & 0.307\\
    &&GPMVS~\cite{hou2019multi} & 16.2 & 7.9 & 12.1 & 0.335 & 0.533 & 0.408\\
    &&DPSNet~\cite{im2019dpsnet} & 17.7 & 8.1 & 12.9 & 0.272 & 0.497 & 0.349\\
    &&DeepVMVS~\cite{duzceker2021deepvideomvs} & 11.7 & 7.6 & 9.7 & 0.451 & 0.558 & 0.496\\
    &&CVD~\cite{luo2020consistent} & 34.4 & 9.1 & 21.8 & 0.266 & 0.461 & 0.331\\
    &&SimRec~\cite{sayed2022simplerecon} & \textbf{\textcolor{blue}{6.5}} & 7.8 & \textbf{\textcolor{blue}{7.2}} & \textbf{\textcolor{blue}{0.641}} & 0.581 & \textbf{\textcolor{blue}{0.608}}\\
    \midrule
    \multirow{6}{*}{\rotatebox{90}{Feature Fusion}}
    &\multirow{3}{*}{\rotatebox{90}{Offline}}&Atlas~\cite{murez2020atlas} & 8.4 & 10.2 & 9.3 & 0.565 & 0.598 & 0.578\\
    &&3DVNet~\cite{rich20213dvnet} & 22.1 & \textbf{\textcolor{teal}{7.7}} & 14.9 & 0.506 & 0.545 & 0.520\\
    &&VoRTX~\cite{stier2021vortx} & \textbf{\textcolor{teal}{6.2}} & 8.2 & \textbf{\textcolor{teal}{7.2}} & \textbf{\textcolor{teal}{0.688}} & \textbf{\textcolor{teal}{0.607}} & \textbf{\textcolor{teal}{0.644}}\\\cmidrule{2-9}
    &\multirow{3}{*}{\rotatebox{90}{Online}}&NeuRec~\cite{sun2021neuralrecon} & \textbf{\textcolor{violet}{5.4}} & 12.8 & 9.1 & 0.684 & 0.479 & 0.562\\ 
    &&TF~\cite{bozic2021transformerfusion} & 7.8 & \textbf{\textcolor{violet}{9.9}} & 8.9 & 0.648 & \textbf{\textcolor{violet}{0.547}} & 0.591\\
    &&Ours & 5.5 & 10.5 & \textbf{\textcolor{violet}{8.0}} & \textbf{\textcolor{violet}{0.695}} & 0.527 & \textbf{\textcolor{violet}{0.598}}\\
    \bottomrule
  \end{tabular}
}
  \vspace{-0.2cm}
  \caption{\textbf{Quantitative results of 3D metrics on ScanNet.} We show the results of two-stage depth fusion methods (top) and those for end-to-end feature fusion works (bottom) following the evaluation protocol in~\cite{sun2021neuralrecon}. We highlight the best results for~\emph{Depth Fusion},~\emph{Feature Fusion Offline} and \emph{Feature Fusion Online} methods in \textcolor{blue}{blue}, \textcolor{teal}{teal}, and \textcolor{violet}{violet}, respectively. Offline methods assume to observe the whole video sequence. Our method performs the best among all online feature fusion methods in the \textit{Chamfer} metric.}
  \vspace{-0.2cm}
  \label{tab:overall_results}
\end{table}
\begin{table}[!t]
  \centering
  \resizebox{\linewidth}{!}{
  \begin{tabular}{@{}ccccccc@{}}
    \toprule
    Method & Acc $\downarrow$ & Comp $\downarrow$ & Chamfer $\downarrow$ & Prec $\uparrow$ & Recall $\uparrow$ & F-score $\uparrow$\\
    \midrule
    SimRec~\cite{sayed2022simplerecon} & 8.0 & 8.6 & 8.3 & 0.511 & 0.482 & 0.495\\
    \midrule
    NeuRec~\cite{sun2021neuralrecon} & 6.1 & 19.4 & 12.8 & 0.588 & 0.347 & 0.431\\
    Ours & \textbf{\textcolor{violet}{5.9}} & \textbf{\textcolor{violet}{13.1}} & \textbf{\textcolor{violet}{9.5}} & \textbf{\textcolor{violet}{0.620}} & \textbf{\textcolor{violet}{0.441}} & \textbf{\textcolor{violet}{0.512}}\\
    \bottomrule
  \end{tabular}
  }
  \vspace{-0.2cm}
  \caption{\textbf{Quantitative results of 3D metrics on 7-scenes.} We evaluate our method on the official test split of 7-scenes. All methods are trained on ScanNet and for baseline methods, we use their released pre-trained models. }
  \label{tab:seven-scene}
  \vspace{-0.2cm}
\end{table}
\begin{table}[!t]
  \centering
  \resizebox{\linewidth}{!}{
  \begin{tabular}{@{}ccccccccccc@{}}
    \toprule
    & && Acc$\downarrow$ & Comp $\downarrow$ & Chamfer $\downarrow$ & Prec $\uparrow$ & Recall $\uparrow$ & F-score $\uparrow$ \\
    \midrule
    i & w/o any && 5.6 & 14.1 & 9.9 & 0.661 & 0.441 & 0.527 \\
    ii & w resi && 5.8 & 12.0 & 8.9 & 0.649 & 0.457 & 0.535 \\
    iii & w loc, resi && 6.7 & 11.0 & 8.9 & 0.629 & 0.477 & 0.541\\
    iv & w vis, resi && \textbf{5.4} & 11.8 & 8.6 & 0.693 & 0.502 & 0.580 \\
    v & ours (top k) && 5.6 & 11.6 & 8.6 & 0.673 & 0.495 & 0.569 \\
    vi & \textbf{ours} (sliding window) && 5.5 & \textbf{10.5} & \textbf{8.0} & \textbf{0.695} & \textbf{0.527} & \textbf{0.598} \\
    \bottomrule
  \end{tabular}
  }
  \vspace{-0.2cm}
  \caption{\textbf{Ablation study.} We ablate our visibility-based feature fusion (``vis''), ray-based local sparsification (``loc'') and TSDF residual learning (``resi'') on ScanNet. }
  \vspace{-0.1cm}
  \label{tab:ablation}
\end{table}
\begin{figure*}[!ht]
  \centering
  \includegraphics[width=0.96\textwidth]{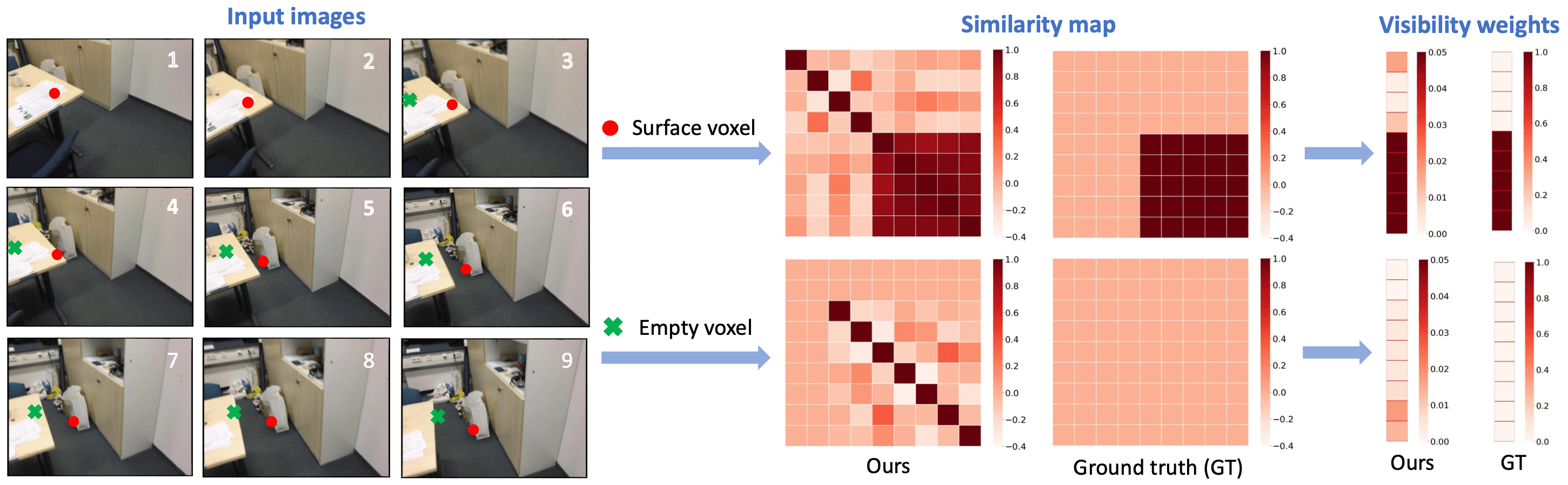}
  \vspace{-0.3cm}
  \caption{\textbf{Relations between the similarity map and the visibility weights.} Here, we illustrate this relationship using two kinds of voxels. The surface voxel (red dot) occupied by the paper bag on the floor is visible in the last 5 views. Thus the features extracted from those images have higher similarity with each other than those for the other views. So do the visibility weights. For the empty voxel (green cross), the features from different images are different leading to lower visibility weights.}
  \label{fig:vis_similarity}
  \vspace{-0.1cm}
\end{figure*}
\begin{figure*}[!t]
    \centering
    \begin{tabular}{cccc}
    SimpleRecon~\cite{sayed2022simplerecon} & NeuralRecon~\cite{sun2021neuralrecon} & Ours & Ground truth \\
    \includegraphics[width=0.225\linewidth]{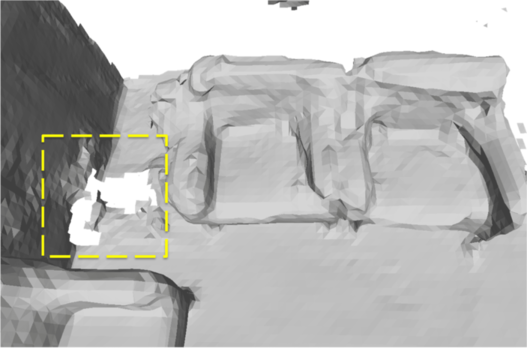} & \includegraphics[width=0.225\linewidth]{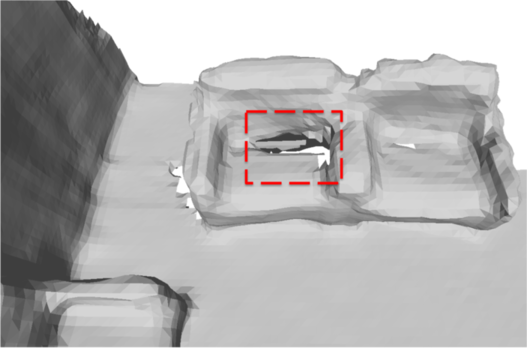} & \includegraphics[width=0.225\linewidth]{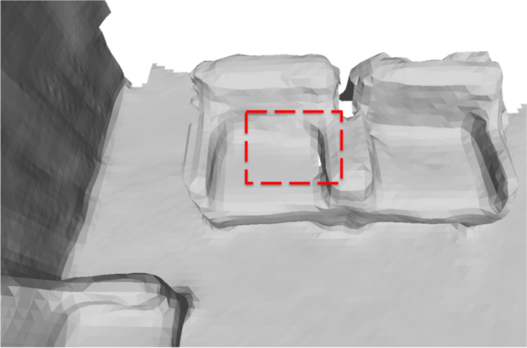} & \includegraphics[width=0.225\linewidth]{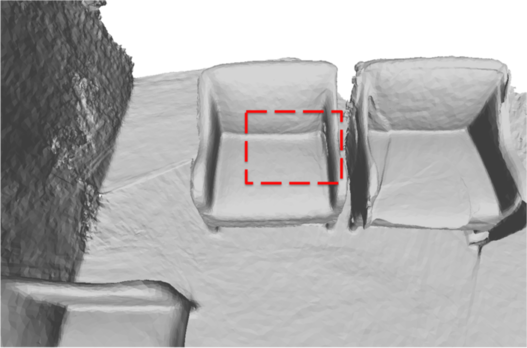}\\
    
    \includegraphics[width=0.225\linewidth]{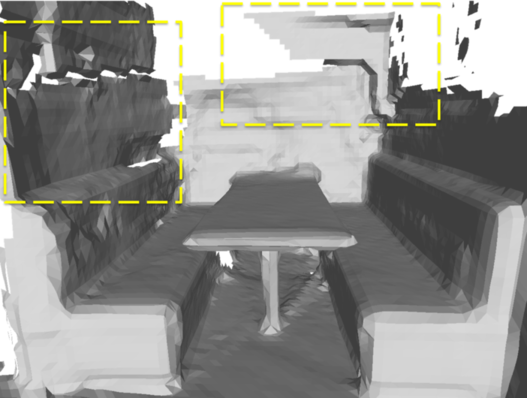} & \includegraphics[width=0.225\linewidth]{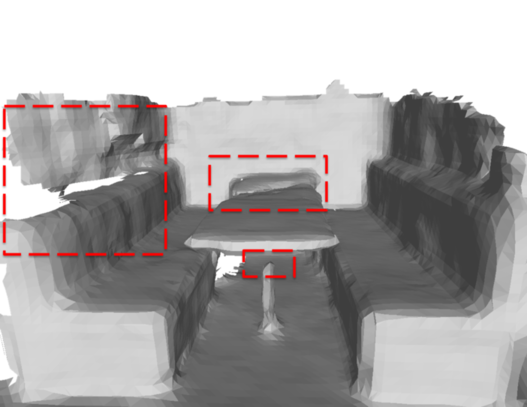} & \includegraphics[width=0.225\linewidth]{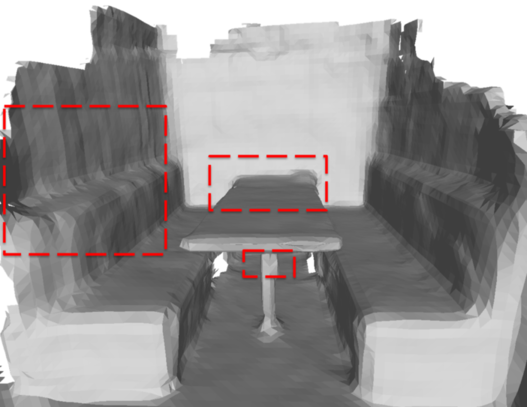} & \includegraphics[width=0.225\linewidth]{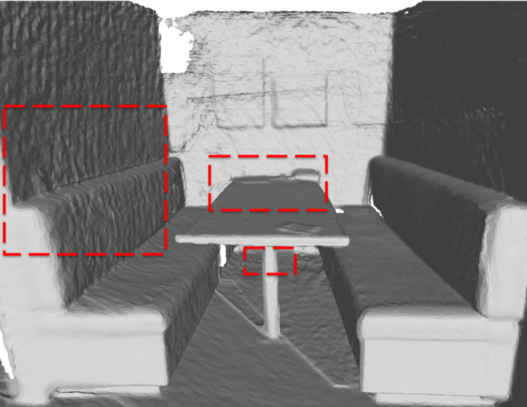}\\
    
    \includegraphics[width=0.225\linewidth]{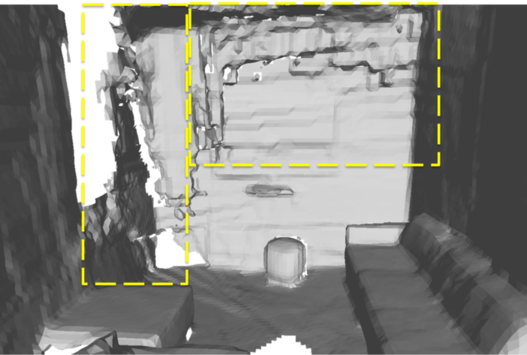} & \includegraphics[width=0.225\linewidth]{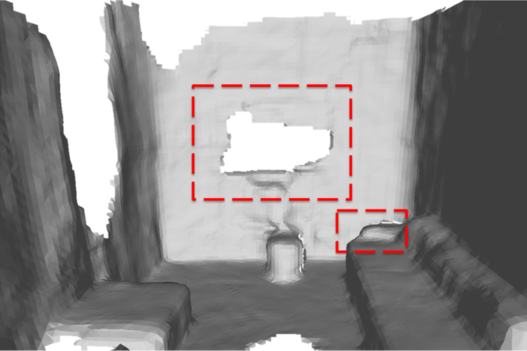} & \includegraphics[width=0.225\linewidth]{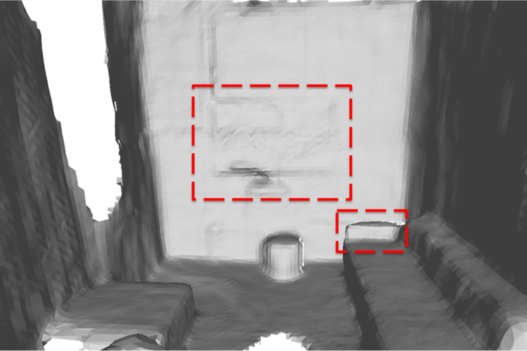} &
    \includegraphics[width=0.225\linewidth]{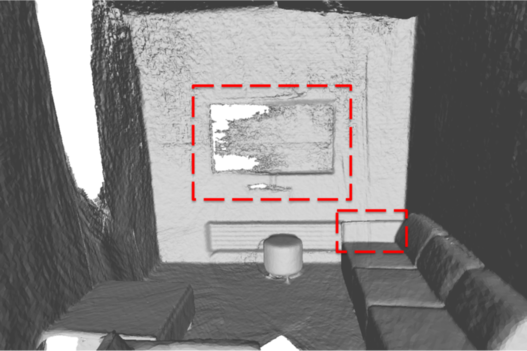}\\
    
    \includegraphics[width=0.225\linewidth]{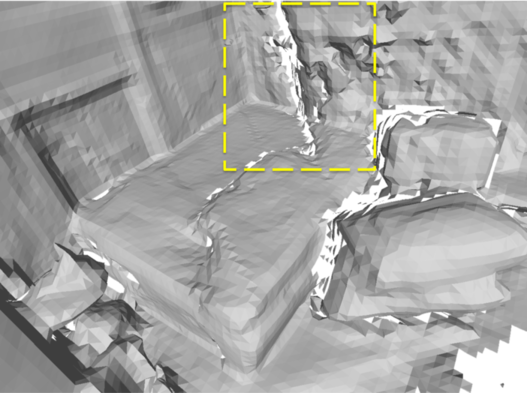} & \includegraphics[width=0.225\linewidth]{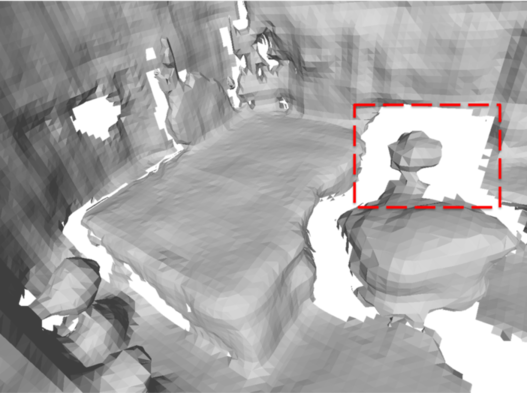} & \includegraphics[width=0.225\linewidth]{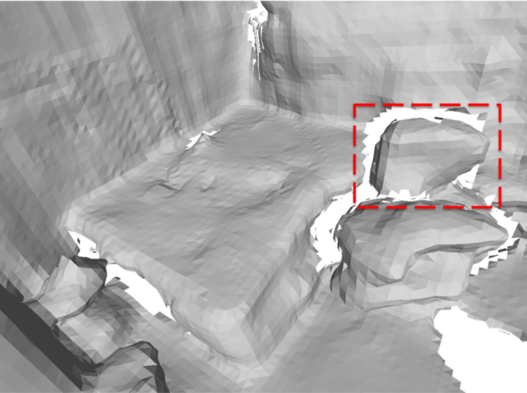} & \includegraphics[width=0.225\linewidth]{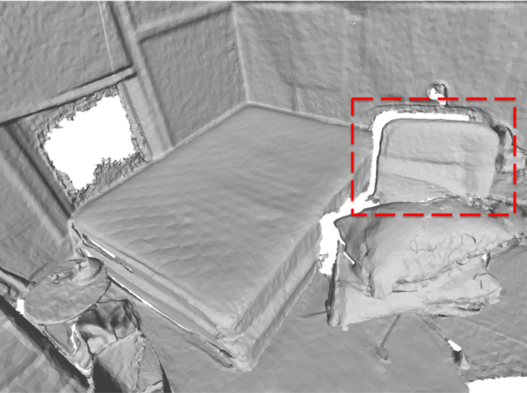}\\

    \includegraphics[width=0.225\linewidth]{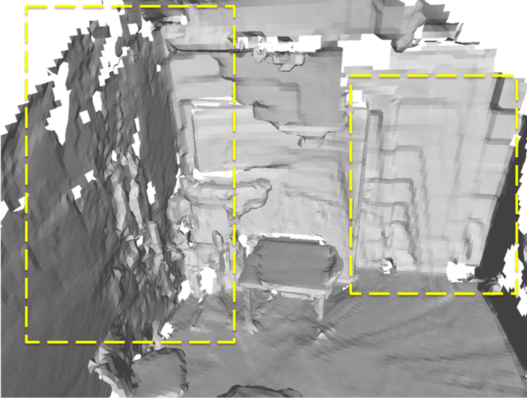} & \includegraphics[width=0.225\linewidth]{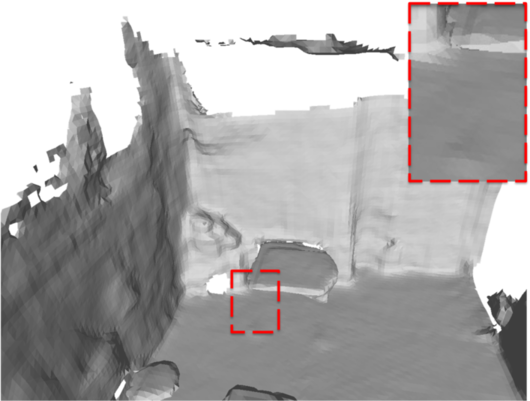} & \includegraphics[width=0.225\linewidth]{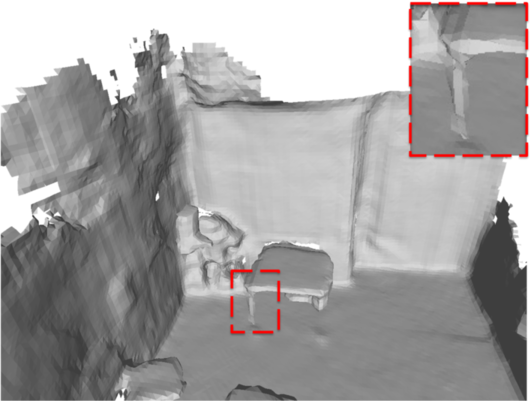} & \includegraphics[width=0.225\linewidth]{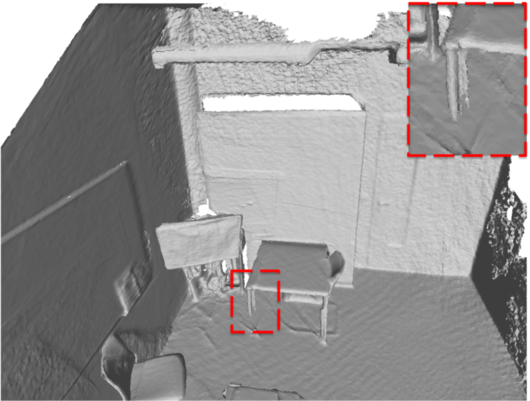}\\
    
    \includegraphics[width=0.225\linewidth]{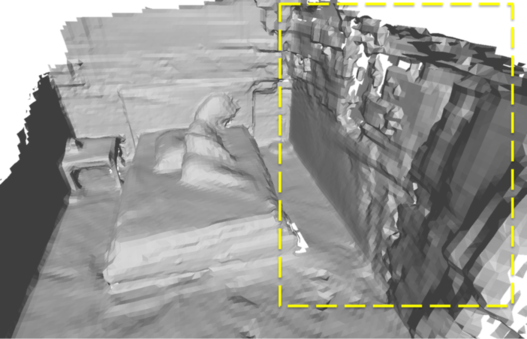} & \includegraphics[width=0.225\linewidth]{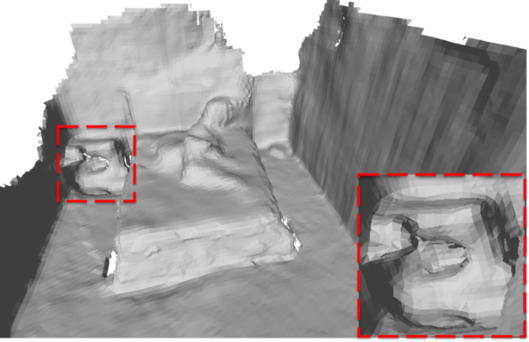} & \includegraphics[width=0.225\linewidth]{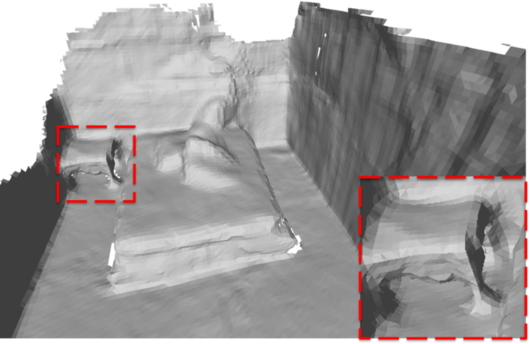} & \includegraphics[width=0.225\linewidth]{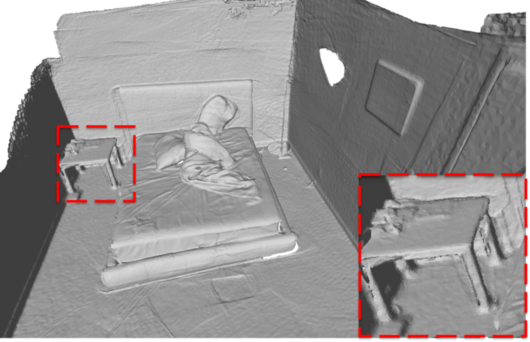}\\
    \end{tabular}
    \vspace{-0.25cm}
    \caption{\textbf{Qualitative comparison on ScanNet.} Compared to NeuralRecon~\cite{sun2021neuralrecon}, our reconstruction results are more complete and contain more details (highlighted in the red boxes). Since SimpleRecon~\cite{sayed2022simplerecon} is a two-stage depth-based method, it generates many artifacts and is not coherent (highlighted in the yellow boxes).}
    \vspace{-0.1cm}
    \label{fig:quali_compare}
\end{figure*}

\subsection{Datasets, Metrics \& Baselines}
\noindent\textbf{Datasets.} We evaluate our method on the ScanNet (V2) ~\cite{dai2017scannet} and 7-Scenes dataset~\cite{shotton2013scene}.~ScanNet consists of 2.5M images in 1613 scans across 807 distinct scenes with ground-truth depths, camera poses, surface reconstructions, and instance-level semantic segmentations. We use the official train/val/test split to train and evaluate our method. For 7-Scenes, it contains 7 different scenes recorded from a handheld Kinect RGB-D camera. Each distinct scene is scanned several times by different users to generate multiple image sequences. We further evaluate our trained model on the official test split of 7-Scenes directly to demonstrate the generalization ability of our method. Details about datasets are included in the supplementary material.

\noindent\textbf{Metrics.} 
We follow the 3D geometry metrics used in~\cite{murez2020atlas, sun2021neuralrecon, bozic2021transformerfusion} to evaluate the performance of our approach. The detailed definitions of these metrics are included in the supplementary material. Among these metrics, we regard the \textit{Chamfer} distance as the most important metric because it is the average of accuracy, measuring the distance from predicted point clouds to the ground truth ones, and completeness, measuring the distance from ground truth point clouds to the predicted ones, for all 3D points. Although \textit{F-score} also represents the balance between accuracy and completeness, it is also affected by a user-defined threshold ($5$cm).
    
\noindent\textbf{Baselines.}
We compare our method with state-of-the-art depth-based methods and end-to-end reconstruction methods. For a fair comparison, these methods are trained and evaluated by following the official data split provided by the ScanNet dataset~\cite{dai2017scannet}. We use the pre-trained model for~\cite{duzceker2021deepvideomvs} and fine-tuned models for~\cite{wang2018mvdepthnet, hou2019multi, im2019dpsnet} provided by~\cite{duzceker2021deepvideomvs}. Following~\cite{sayed2022simplerecon}, the maximum fused depth is limited to 3m. The results of~\cite{sayed2022simplerecon, murez2020atlas, rich20213dvnet, stier2021vortx, bozic2021transformerfusion} are taken from~\cite{sayed2022simplerecon} and~\cite{schonberger2016pixelwise, luo2020consistent, sun2021neuralrecon} are taken from~\cite{sun2021neuralrecon}. Among depth-based methods, MVDepthNet~\cite{wang2018mvdepthnet}, GPMVS~\cite{hou2019multi} and DeepVideoMVS~\cite{duzceker2021deepvideomvs} achieve real-time performance while GPMVS, DeepVideoMVS and CVD~\cite{luo2020consistent} are depth-fusion baselines with consistent video depth estimation. NeuralRecon~\cite{sun2021neuralrecon} and TransformerFusion~\cite{bozic2021transformerfusion} are two end-to-end incremental volumetric reconstruction frameworks that directly predict the surface geometry and are the most relevant ones to our approach.

\subsection{Results}
\noindent\textbf{ScanNet.}
The experimental results on the ScanNet dataset are reported in \cref{tab:overall_results}. Our method outperforms all existing online feature fusion methods~\cite{sun2021neuralrecon, bozic2021transformerfusion} in both \textit{Chamfer} distance and \textit{F-score} metrics. We achieve a $12.1\%$ reduction in chamfer distance compared to NeuralRecon~\cite{sun2021neuralrecon}. Although the concurrent work SimpleRecon~\cite{sayed2022simplerecon} achieves slightly better numerical results, due to their two-stage pipeline, their results often include noises especially for low texture regions like walls as shown in~\cref{fig:quali_compare}. In contrast, our reconstruction results are more clean and coherent. We also compare our qualitative results to those of NeuralRecon~\cite{sun2021neuralrecon} in~\cref{fig:quali_compare}. As highlighted in red boxes, our visibility-aware feature fusion and ray-based sparsification enable us to reconstruct more complete and detailed scene structures, demonstrating the effectiveness of our approach. 

\noindent\textbf{7-Scenes.} We also report the results on 7-Scenes in~\cref{tab:seven-scene}. The conclusion still holds. As the data split used by~\cite{sun2021neuralrecon} is not available, we thus report the results of evaluating their pretrained model on the official test split of 7-Scenes. It may explain why the results of~\cite{sun2021neuralrecon} are different from those reported in the original paper. More results on 7-Scenes are provided in the supplementary material.

\noindent\textbf{Visibilities.} In~\cref{fig:vis_similarity}, we visualise the similarity map and learned visibility weights for surface voxels and empty voxels in the current local fragment, respectively. The similarity map shows that the photometric consistency between features from different views provides strong heuristics for voxel occupancy probability and view selection. From the visibility weights, we can find our local feature fusion module is able to select the most relevant views for each voxel to local occupancy prediction. More visualisations are shown in the supplementary material.

\noindent\textbf{Ablation study.}
To demonstrate the effectiveness of each module of our method, we conduct several experiments on the ScanNet dataset and show the results in \cref{tab:ablation}. Comparing row i and row ii, by adding the residual learning strategy, the chamfer distance is reduced by $10.1\%$. As shown in row iii and row iv, with our visibility predicting module and ray-based local sparsification strategy, our performance is further improved. Finally, we compare the effectiveness of using a sliding window to select consecutive $K$ voxels, to simply choosing the top $K$ voxels, along each ray in row v and row vi. It shows that utilising the sliding window for sparsification improves the performance.

\noindent\textbf{Running time.}
Although due to the extra computation of our visibility-aware feature fusion and ray-based local sparsification, our method tends to be slower than neuralrecon~\cite{sun2021neuralrecon}, our model still achieves a real-time reconstruction of 25 key frames per second (FPS) on an NVIDIA RTX 2080Ti GPU and 45 FPS on an NVIDIA RTX 4090 GPU.

\section{Conclusion}
\label{sec:conclusion}

In this paper, we demonstrate that explicitly learning the visibility weights for feature fusion and adopting a ray-based local sparsification strategy will benefit the online incremental 3D scene reconstruction task. In the future, we will try to combine the advantages of depth estimation and end-to-end feature-based reconstruction to generate both sharper and coherent surface geometry.

\noindent{\textbf{Acknowledgements}}

This research was supported in part by the Australia Research Council DECRA Fellowship (DE180100628) and ARC Discovery Grant (DP200102274). 

{\small
\bibliographystyle{ieee_fullname}
\bibliography{egbib}
}

\end{document}